\documentclass[preprint,12pt]{larticle}
\usepackage[a4paper, total={6.18in, 10in}]{geometry}

\usepackage{iftex}
\ifPDFTeX
  \usepackage[utf8]{inputenc}
\else
  \usepackage{fontspec}
  \usepackage{xeCJK}
  \IfFontExistsTF{Noto Serif CJK SC}
    {
      \setCJKmainfont{Noto Serif CJK SC}
      \setCJKsansfont{Noto Sans CJK SC}
      \setCJKmonofont{Noto Sans Mono CJK SC}
    }
    {
      \setCJKmainfont[Script=Default]{FandolSong-Regular.otf}
      \setCJKsansfont[Script=Default]{FandolHei-Regular.otf}
      \setCJKmonofont[Script=Default]{FandolFang-Regular.otf}
    }
  \IfFontExistsTF{UnBatang.ttf}
    {
      \setCJKfallbackfamilyfont{\CJKrmdefault}[Script=Default]{UnBatang.ttf}
      \xeCJKenablefallback
    }
    {}
\fi
\usepackage{color}
\usepackage[dvipsnames]{xcolor}

\usepackage{amssymb}
\usepackage{amsthm}
\usepackage{amsmath}
\usepackage{mathpazo}

\usepackage[symbol]{footmisc}

\newcommand{\Var}{\operatorname{var}}
\newcommand{\T}[1]{\textit{#1}}

\begin{document}

\begin{frontmatter}

\title{Measuring language complexity from hierarchical reuse of recurring patterns}

\author{
Junyi Zhou{\footnotesize{$^{1,2,5\dagger}$}}, Rui Liu{\footnotesize{$^{3,4\dagger}$}},
Pengyu Liu{\footnotesize{$^{6,7*}$}}, Yu Liu{\footnotesize{$^{1,2*}$}}
}
\nonumnote{$^{\dagger}$These authors contributed equally to this work. $^{*}$Co-corresponding authors who contributed equally to this work; correspondence should be addressed to Pengyu Liu (pengyu.liu@uri.edu) or Yu~Liu (yu.ernest.liu@bnu.edu.cn)}

\address{\footnotesize{$^1$}Department of Systems Science, Faculty of Arts and Sciences,
\footnotesize{$^2$}International Academic Center of Complex Systems, 
\footnotesize{$^3$}Department of Chinese Language and Literature, Faculty of Arts and Sciences, and \footnotesize{$^4$}Center for Linguistic Sciences, Beijing Normal University, Zhuhai 519087, Guangdong, China. \\

\footnotesize{$^5$}School of Systems Science, Beijing Normal University, Beijing 100875, China. \\

\footnotesize{$^6$}Department of Mathematics and Applied Mathematical Sciences and \footnotesize{$^7$}Department of Cell and Molecular Biology, University of Rhode Island, Kingston, RI 02881, USA.
}

\begin{abstract}
We introduce the ladderpath index as a measure of language complexity grounded in algorithmic information theory. 
It counts the minimum steps needed to reconstruct a sequence through hierarchical reuse of repeated substructures, capturing an exactly computable but constrained form of algorithmic compressibility related to, but distinct from, Kolmogorov complexity.
We apply the ladderpath approach to 21 parallel corpora from the Parallel Universal Dependencies dataset. 
The ladderpath index is approximately invariant across the languages, and varies much less than the corpus length.
This is more pronounced when all corpora are mapped to a unified binary representation, providing evidence for the equi-complexity hypothesis from a representation-independent perspective.
We also observe trade-offs between character inventory size and corpus length, and between vocabulary-level and corpus-level reconstruction complexity, supporting the trade-off hypothesis that total complexity is conserved and redistributed across linguistic levels.
The reusable substructures identified by the ladderpath approach, without any linguistic input, overlap with words and morphological components attested in the natural vocabulary. 
The hierarchical reuse captured by the ladderpath approach parallels the chunking mechanisms proposed in cognitive science, where the human cognitive system compresses linguistic input into nested, reusable units under shared memory and processing constraints.
This connection between cognitive chunking and the ladderpath approach provides a new interpretation for the equi-complexity and trade-off hypotheses, grounding both in the shared cognitive architecture that underlies language processing across human languages.
\end{abstract}

\end{frontmatter}


\section*{Introduction}
\label{S:1}

Measuring the complexity of natural language is a long-standing challenge in linguistics, in part due to its multifaceted and multi-level nature \cite{ChristianBentz23LinguisticsVanguard}. 
Language encodes information at the lexical, morphological, syntactic, and phonological levels, and complexities at different levels can differ \cite{Housen2019}. 
There is no single measure that can comprehensively quantify language complexity, and the choice of measure often depends on the aspect of investigation \cite{Ortega2012}.
Approaches to quantifying language complexity can be broadly organized into two categories: feature-based linguistic measures and information-theoretic measures.

Feature-based linguistic measures target specific, well-defined structural properties of language at different levels.
Common syntactic-level measures include mean dependency distance (MDD), which reflects the average linear span between syntactically related word pairs within sentences \cite{Liu2008DependencyDA, Liu17PhysLifeRev}, mean length of T-unit (MLT), which quantifies the average length of minimal independent syntactic units \cite{Ortega2003, Liu2011TESOL}, and global distances between dependency structures of sentences \cite{liu2022dptree}.
Morphological-level indices include the index of synthesis, that is the mean number of morphemes per word \cite{Coltekin2023}, and paradigm entropy, which measures the unpredictability of inflectional forms within a language's paradigm system \cite{Cotterell2019}. 
At the lexical level, measures such as the type-token ratio (TTR) and the measure of textual lexical diversity (MTLD) are designed to quantify the range and richness of vocabulary of a text \cite{Jarvis2013}.

Information-theoretic approaches characterize complexity at the level of the whole text rather than through isolated linguistic features.
Shannon entropy, applied to the symbol distributions of natural language, provides a measure of the average information content per symbol and has been used to estimate the entropy rate across multiple corpora and languages \cite{Shannon48BellSystem, Koplenig23SciRep}.
However, Shannon entropy is sensitive only to the frequency distribution of symbols and does not directly account for structural complexity arising from repeated substructures within a text. 
Measures based on Kolmogorov complexity address this limitation, as they define the complexity of a text as the length of the shortest program capable of reconstructing it---a quantity that, in principle, captures redundancy introduced by recurring patterns at multiple scales \cite{Kolmogorov1968, Ehret2021, Liu2024Humanities}.
In practice, Kolmogorov complexity is not accurately computable, and can only be approximated, typically through compression algorithms \cite{Vitanyi2020}.

In this paper, we focus on one computable source of algorithmic compressibility: the reuse of repeated substructures within a sequence.
The ladderpath approach \cite{Liu2022Entropy, Liu2024PRR} formalizes this idea by reconstructing a target sequence from basic units through a shortest hierarchy of reusable subsequences.
The resulting \textit{ladderpath index} ($\lambda$) is the minimum number of reconstruction steps under this constrained scheme; we use it as a measure of language complexity to compare parallel corpora across languages.

\section*{Results}
\label{S:2}

\subsection*{Ladderpath index as a measure of language complexity.}

The ladderpath approach is a method for decomposing a sequence into hierarchical structures built from repetitive elements. A detailed description of the ladderpath approach can be found in \cite{Liu2022Entropy, Liu2024npj, Liu2024PepHiRe}; only a brief recap is provided here. 
Given a target sequence and a set of basic building blocks (the smallest non-separable elements at the chosen level of analysis), the ladderpath approach seeks the most efficient way to reconstruct the target sequence by combining basic building blocks into intermediate building blocks, combining intermediate ones into larger ones, and reusing any constructed blocks wherever they recur.
These intermediate reusable building blocks are called \textit{ladderons}: recurring subsequences that are constructed once and reused at least once during reconstruction. Each ladderon and each basic building block is associated with a \textit{multiplicity}, defined as the number of times that is reused during the reconstruction process. 
The ladderons and basic building blocks together form a hierarchically organized structure called a \textit{laddergraph}, which encodes the nested compositional relationships among all reused components.

The reconstruction proceeds through a sequence of generation-operations. Each generation-operation concatenates two previously introduced or constructed subsequences (basic building blocks or ladderons) to form a longer subsequence.
Once a subsequence has been constructed, it is available for reuse in subsequent generation-operations: each reuse counts as a single generation-operation regardless of the length of the subsequence being reused.
The shortest path, namely the \textit{ladderpath}, of a target sequence is the minimal chain of generation-operations needed to reconstruct it from basic building blocks under this reuse principle. 
Because reuse eliminates the need to independently reconstruct identical subsequences, target sequences with more internal repetition require fewer generation-operations relative to their length.

\begin{figure}[ht!]
    \centering
    \includegraphics[width=1.0\linewidth]{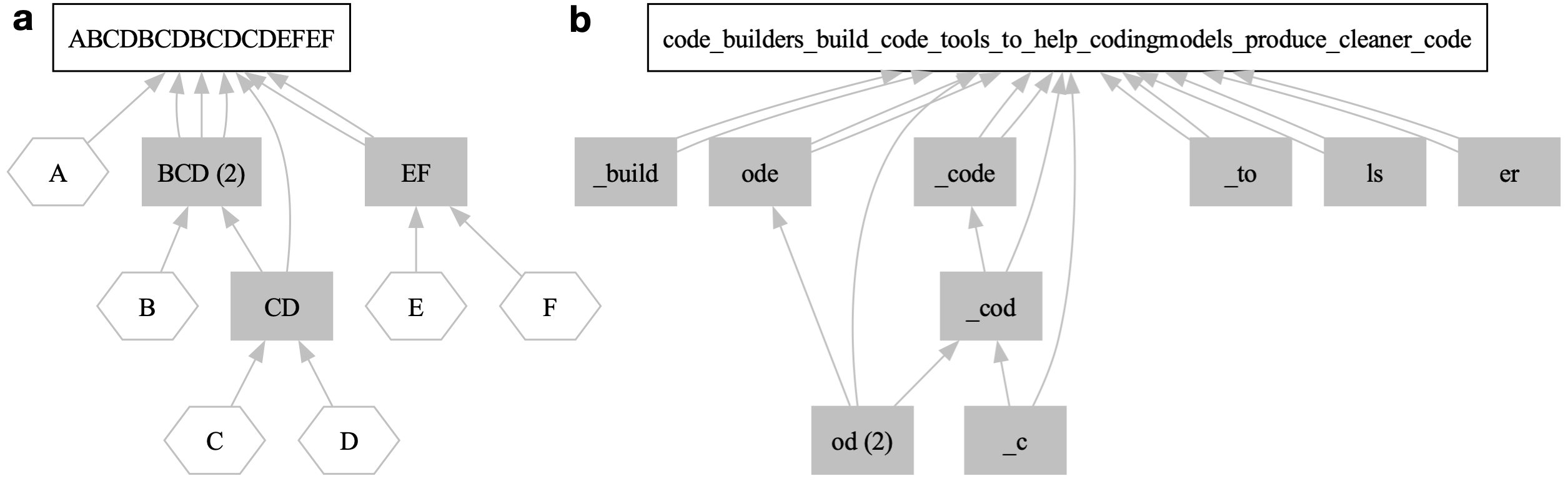}
    \caption{\textbf{Ladderpath decomposition of a sequence.} (a) A toy example illustrates how the ladderpath approach reconstructs a target sequence by composing basic building blocks into reusable subsequences. White nodes denote basic building blocks, gray boxes denote ladderons, and arrows indicate compositional relations used to build the target sequence. The multiplicity of each block is shown in parentheses after the block label; multiplicities equal to 1 are omitted for visual clarity. The size index $S$ counts the reconstruction cost without reuse, whereas the ladderpath index $\lambda$ counts the minimum reconstruction cost when repeated subsequences can be reused hierarchically. (b) A natural-language example shows ladderons extracted from repeated character-level subsequences. For visual clarity, the basic building blocks in panel (b) are omitted.}
    \label{fig1}
\end{figure}

Two complexity measures emerge from the ladderpath approach, which seeks the shortest path for reconstructing a sequence.
The ladderpath index ($\lambda$) is the length of this shortest path, that is, the minimal number of generation-operations needed to reconstruct the target sequence.
Because this quantity measures the operational cost of reconstructing a sequence from basic building blocks through hierarchical reuse, we also refer to it as the {\em reconstruction cost} and use the two terms interchangeably throughout this paper.
The \textit{size index} ($S$) is the length of the trivial reconstruction process, where no reuse is permitted; for a character string, this equals the number of characters.
For example, consider the target sequence ``ABCDBCDBCDCDEFEF'' analyzed at the character level, where each distinct character is a basic building block (Fig. \ref{fig1}a).
The sequence has 16 characters, so a trivial reconstruction with no reuse requires 16 generation-operations and gives $S=16$.
This sequence contains repeated subsequences that can be reused.
In the ladderpath decomposition, such shared subsequences are ladderons, which are constructed once and then reused in later steps.
For instance, ``CD'' is constructed from its constituent characters and reused when building both ``BCD'' and the target sequence. Because ``CD'' is reused once after it is constructed, its multiplicity is 1 (in Fig. \ref{fig1}a, multiplicities equal to 1 are omitted for visual simplicity); By contrast, ``BCD'' is used three times in reconstructing the target sequence, and is therefore reused twice after its construction, giving it a multiplicity of 2, which is shown in parentheses in the figure. 
The result is a nested hierarchy: characters compose into short ladderons, short ladderons compose into longer ladderons, and these components ultimately reconstruct the target sequence.

In Fig. \ref{fig1}a, the target sequence has ladderpath index $\lambda=10$.
First, three operations construct the reusable subsequences: ``C'' + ``D'' gives ``CD'', ``B'' + ``CD'' gives ``BCD'', and ``E'' + ``F'' gives ``EF''. 
Here, ``CD'' can be directly reused when constructing ``BCD'' because it has already been constructed in an earlier step. 
Next, six operations concatenate the seven components ``A'', ``BCD'', ``BCD'', ``BCD'', ``CD'', ``EF'', and ``EF''. 
One final operation then retrieves the completed target sequence, giving $3+6+1=10$ operations in total.
Additionally, Fig. \ref{fig1}b shows the same analysis for a natural-language sequence, where the size index is $S=73$ and the ladderpath index is $\lambda=52$.

The ladderpath index bears a conceptual relationship to Kolmogorov complexity, which measures the length of the shortest computer program that produces a given string as output \cite{Kolmogorov65PIT}.
Both quantities address a related underlying question about how concisely an object can be described, but they differ in scope: Kolmogorov complexity captures any computable regularity, including symmetries, arithmetic progressions, and recursive rules, whereas the ladderpath index specifically captures the reduction achievable through the reuse of repeated substructures.
Another fundamental distinction is that Kolmogorov complexity is uncomputable, that is, no algorithm can determine its exact value for an arbitrary string \cite{Vitanyi2020, Liu2021SciAdv}. 
In practice, Kolmogorov complexity can only be approximated, for instance through compression algorithms such as Lempel-Ziv \cite{Ziv1977}. 
The ladderpath index, by contrast, is exactly computable. 
Although the computation is NP-hard in general, algorithms have been developed and applied to sequences of practical length \cite{Liu2024PRR, Liu2025Alg}.
The ladderpath index thus provides a computable, deterministic measure that captures a specific aspect of what Kolmogorov complexity measures in principle: the reduction in description length achievable through the reuse of repeated substructures.

The hierarchical decomposition produced by the ladderpath approach has a natural interpretation in cognitive science.
A foundational concept in cognitive science is chunking \cite{Miller1956}, where the cognitive system groups individual items into larger units to overcome the limited capacity of short-term memory. 
This concept has been extended into a model of language processing \cite{Christiansen2016}. 
Because memory for linguistic input is extremely transient, the cognitive system must rapidly compress incoming material into increasingly abstract levels of representation.
At the morphological level, frequently co-occurring character sequences are consolidated into recognized morphemes and word forms. 
At the syntactic level, words that regularly appear together are grouped into fixed phrases, collocations, and constructional patterns that function as single processing units. 
For example, a phrase such as ``in front of'' is not parsed word by word but retrieved as a single prepositional chunk. 
These cognitive chunks form a nested hierarchy.
Smaller chunks serve as building blocks for larger ones, and all are stored in memory for rapid retrieval and reuse during language comprehension and production.

The ladderons identified by the ladderpath approach are analogous to these cognitive chunks. 
They are reusable substructures at multiple levels of granularity that compose hierarchically into larger units. 
In the context of natural language, the ladderpath index of a corpus measures the minimal number of construction steps needed to generate that corpus from its basic building blocks (e.g., characters or words), given that any previously constructed subsequence can be freely reused.
A corpus with a low ladderpath index relative to its size index contains substantial internal reuse of substructures, such as word stems, affixes, and longer phrases, that reduce the number of independent construction steps.

\subsection*{Consistent ladderpath indices across languages.}

The equi-complexity hypothesis in linguistics states that all human languages are roughly equally complex overall \cite{hockett1958}.
This hypothesis has been widely assumed but difficult to test, in part because there has been no generally accepted definition of overall language complexity.
One approach is to characterize the complexity of a language not by a single metric but by a vector of measurements across multiple linguistic subsystems, and to compare these vectors across languages.
Research based on this approach found that, while an individual metric in a specific subsystem can vary, the overall complexity vectors are largely consistent, which support the equi-complexity hypothesis \cite{Bentz22LingVan}.

\begin{figure}[ht!]
    \centering
    \includegraphics[width=1.0\linewidth]{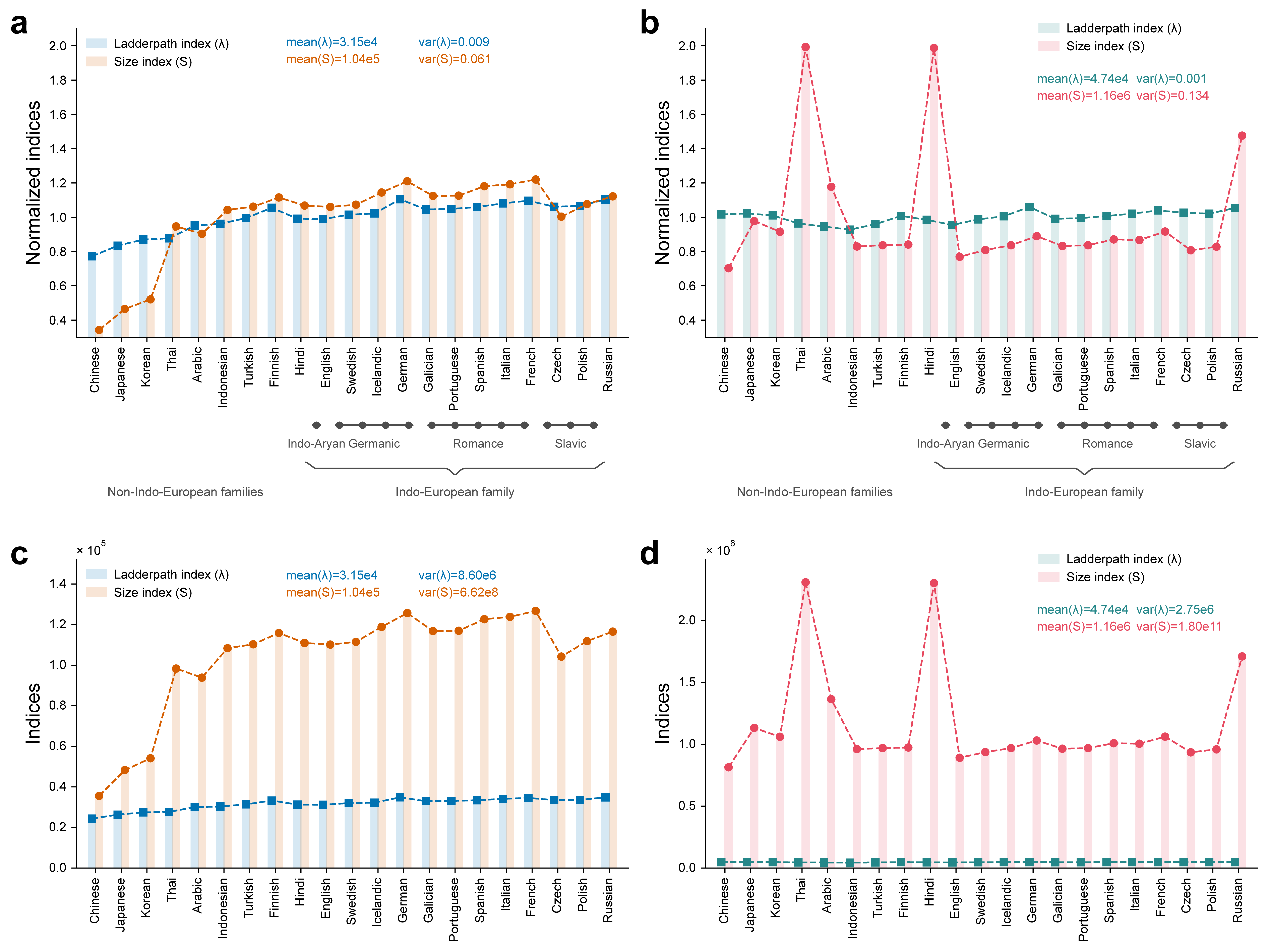}
    \caption{\textbf{Cross-language convergence of the ladderpath index.} (a) Normalized ladderpath index $\lambda$ and size index $S$ for the 21 PUD corpora at the character level. (b) Normalized $\lambda$ and $S$ after all corpora are converted to UTF-8 binary representations. (c) Absolute values of $\lambda$ and $S$ for the character-level analysis in panel (a). (d) Absolute values of $\lambda$ and $S$ for the UTF-8 binary analysis in panel (b). Across both representations, the ladderpath index varies much less across languages than the size index.}
    \label{fig2}
\end{figure}

Our results provide new evidence in support of the equi-complexity hypothesis from the perspective of algorithmic information theory. 
We apply the ladderpath approach to the 21 PUD corpora at the character level. 
The normalized ladderpath index shows much lower cross-language variance than the normalized size index $(\Var(\lambda)=0.009$ vs. $\Var(S)=0.061)$ (Fig. \ref{fig2}a), with the corresponding absolute values shown in Fig. \ref{fig2}c.
This indicates that although the parallel corpora differ substantially in character length, the reconstruction cost from character-level building blocks remains comparatively stable across languages. 
Mild deviations in the ladderpath index are still observed for languages such as Chinese, Japanese and Korean, which likely reflect differences in the character collections of languages. 
To diminish the effect of these differences, we convert each of the 21 parallel corpora into binary representations through UTF-8 encoding. 
The pattern becomes even more pronounced when the analysis is conducted at the bit-level.
The variance of the normalized ladderpath index drops to $\Var(\lambda)=0.001$, whereas the variance of the normalized size index increases to $\Var(S)=0.134$ (Fig. \ref{fig2}b), with the corresponding absolute values shown in Fig. \ref{fig2}d. 
Thus, placing all languages in a common binary representation further reduces cross-language variation in reconstruction cost, suggesting that much of the apparent difference at the character level reflects writing-system granularity rather than a large difference in underlying reconstructive complexity.
We repeat the same analysis on an additional set of parallel corpora and observe the same pattern. See Supplementary Information (SI) Section S1 and S2 for details.

These findings provide evidence for the equi-complexity hypothesis from a new perspective: While languages differ in the corpus length and in the size of their character collections, the complexity of reconstructing each corpus through hierarchical reuse of repeated substructures, as measured by the ladderpath index, is approximately invariant.
From a cognitive science perspective, this is interpretable in terms of the shared processing constraints of the human cognition.
If all human languages must be processed by the same cognitive architecture with the same memory and processing limitations \cite{Christiansen2016}, then it is plausible that the degree of hierarchical reuse in language would be constrained to a similar range.
A language with too little reuse would require speakers to construct most expressions from basic building blocks at the time of use, placing excessive demands on working memory and real-time processing; such a language would be difficult to produce and comprehend within the tight temporal constraints of natural communication.
Conversely, a language with too much reuse would rely heavily on a large collection of prefabricated building blocks in long-term memory, reducing the system's ability to generate novel expressions and limiting its compositional flexibility.
The near-invariance of the ladderpath index across languages may thus reflect a universal cognitive constraint that balances the working memory cost of real-time assembly against the long-term memory cost of storing reusable units, operating independently of the particular structural strategies each language employs.

\subsection*{Ladderpath index trade-off between corpus and vocabulary.}

Closely related to the equi-complexity hypothesis is the trade-off hypothesis, which states that complexity in one domain of language is compensated by simplicity in another. 
Cross-linguistic trade-offs between various linguistic subsystems have been documented, including morphology and syntax \cite{Sinnemaki2014}, word order and word structure \cite{Koplenig2017}, and grammar and vocabulary size \cite{Reali2018}. 
In \cite{Bentz22LingVan}, the authors found that approximately one third of significant correlations between their 28 complexity metrics were negative, providing partial support for the trade-off hypothesis.
The general principle underlying these trade-offs has been linked to Zipf's principle of least effort: if one linguistic subsystem already encodes certain information, it would constitute unnecessary cognitive effort to redundantly encode the same information in another subsystem \cite{Koplenig2017}.

\begin{figure}[ht!]
    \centering
    \includegraphics[width=1.0\linewidth]{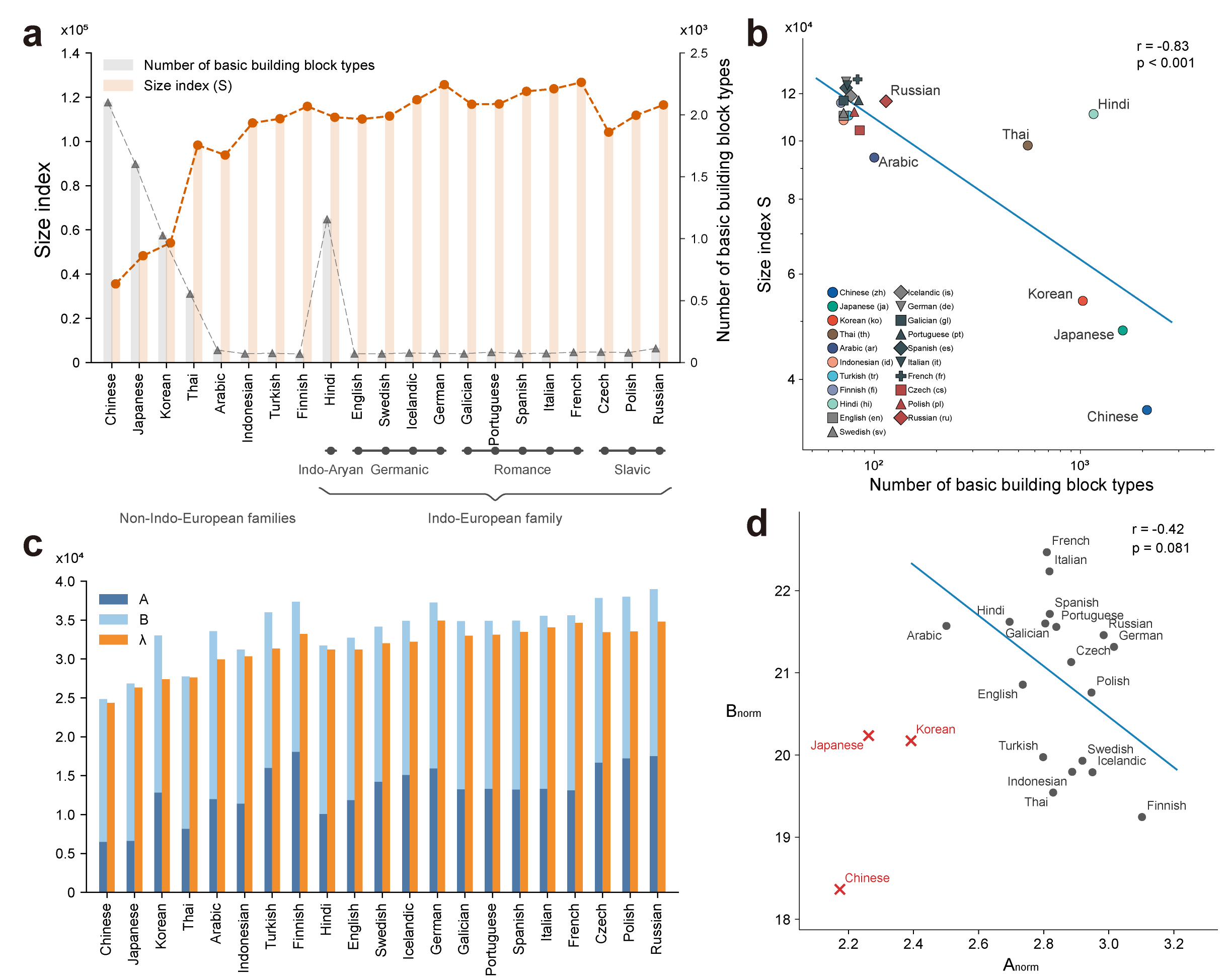}
    \caption{\textbf{Trade-offs between character inventory, corpus length, and reconstruction cost.} (a) Size index $S$ and the number of basic building block types for the 21 PUD corpora at the character level. (b) Negative correlation between $S$ and the number of basic building block types across languages. (c) The constrained two-stage cost $A+B$ is close to, but never below, the direct character-level ladderpath index $\lambda$. (d) Normalized costs $A_{\mathrm{norm}}$ and $B_{\mathrm{norm}}$ show a negative correlation after excluding Chinese, Japanese, and Korean ($r=-0.42$, $p=0.081$).}
    \label{fig3}
\end{figure}

Our results reveal two trade-offs that are naturally expressed by the ladderpath approach.
The first is a trade-off between the size index (namely, corpus length) and the number of basic building block types (Fig. \ref{fig3}a).
Across the 21 PUD corpora, languages with larger character inventories (namely, larger number of basic building block types) tend to have smaller size indices, showing a strong negative Pearson correlation ($r=-0.83$, $p<0.001$; Fig. \ref{fig3}b).
Languages such as Chinese, Japanese, and Korean employ large character collections (over 1,000 distinct character types in the PUD corpora) and produce comparatively short corpora as measured by character count. 
Conversely, alphabetic languages such as English, German, and the Romance languages use small character inventories (typically fewer than 100 distinct characters) and require longer character sequences to express the same content. 
Hindi presents a partial exception, combining a moderately large character inventory with a corpus length comparable to that of alphabetic languages.
This inverse relationship between character inventory size and corpus length is consistent with the general principle that a larger set of basic building blocks allows each unit to carry more information, thereby requiring fewer units to encode the same content.

The second trade-off emerges from a two-stage decomposition of the ladderpath index. For each language, we first take the natural vocabulary defined a priori by the PUD word segmentation as a set of target sequences and compute the ladderpath index from characters to this prior vocabulary; we denote this vocabulary-level ladderpath index by $A$ (Fig. \ref{fig3}c). We then treat the same prior vocabulary as the set of basic building blocks and compute the ladderpath index needed to reconstruct the PUD corpus from these word-level units; we denote this corpus-level ladderpath index by $B$.
The technical procedure for this decomposition is described in the {\em Experiments} subsection of Methods and Materials, with further details provided in SI Section S3.
Together, $A$ and $B$ define a constrained two-stage reconstruction path from characters to the corpus through the prior vocabulary. By contrast, the direct character-level ladderpath index $\lambda$ reconstructs the corpus directly from character-level building blocks, without requiring the path to pass through the prior word segmentation. Because this direct character-level ladderpath is the shortest unconstrained reconstruction path, the two-stage cost should satisfy $A+B \geq \lambda$. As shown in Fig. \ref{fig3}c, this inequality holds across languages, but the gap between $A+B$ and $\lambda$ is generally small. This supports the idea that the prior vocabulary provides an effective intermediate representation: many word-level units supplied by prior segmentation are close to the reusable subsequences that the ladderpath decomposition itself would identify.

We next examine how the two stage-specific ladderpath indices vary across languages. To make these two costs comparable across languages, we normalize $A$ by the number of prior word types in the natural vocabulary, giving $A_{\mathrm{norm}}$, and normalize $B$ by the number of sentences in the corpus, giving $B_{\mathrm{norm}}$ (since each PUD corpus contains 1,000 sentences, the normalization denominator for $B$ is always 1,000 across languages). Thus, $A_{\mathrm{norm}}$ approximates the average cost of constructing a word from characters, whereas $B_{\mathrm{norm}}$ approximates the average cost of assembling a sentence from word-level units.
Because Chinese, Japanese, and Korean have much larger character inventories than the other languages in the character-level representation, they constitute a distinct writing-system regime in this analysis. We therefore examine the correlation primarily among the remaining languages, where writing-system granularity is more comparable.
In this subset, $A_{\mathrm{norm}}$ and $B_{\mathrm{norm}}$ show a negative Pearson correlation ($r=-0.42$, $p=0.081$; Fig. \ref{fig3}d). The negative trend suggests a redistribution of reconstruction effort across levels. Languages with a higher character-to-vocabulary reconstruction cost tend to have a lower vocabulary-to-corpus reconstruction cost: more effort is invested in constructing word-level units, but these units then serve as larger reusable building blocks for reconstructing the corpus. Conversely, languages with a lower vocabulary-level reconstruction cost tend to require more reconstruction effort at the corpus level, because the corpus must be assembled from less internally complex word-level units. This pattern suggests that reconstruction effort is not simply accumulated across levels, but redistributed between constructing word-level units and constructing the corpus from those units. For completeness, we also report the correlation with Chinese, Japanese, and Korean included in SI Section S3. In that case, the correlation becomes weakly positive and non-significant ($r=0.23$, $p=0.307$), suggesting that pooling languages with substantially different writing-system regimes can obscure the within-group trade-off.

From a cognitive science perspective, these trade-offs can be interpreted as reflecting the allocation of processing resources across different levels of linguistic structure. 
As discussed in previous sections, the cognitive system faces a balance between working memory demands during real-time assembly and long-term memory demands for storing reusable building blocks. 
A language with a large character inventory, such as Chinese, encodes more information per character, requiring fewer characters to express the same content and thus producing a shorter corpus. 
However, this places a greater demand on long-term memory, as speakers and readers must learn and retain a larger set of basic symbols. 
A language with a small character inventory, such as English, encodes less information per character and requires longer character sequences to convey the same content, but imposes a lighter long-term memory burden for learning the basic symbol set. 
The cognitive cost is redistributed rather than eliminated: what is saved in long-term memory for symbol storage is paid in the processing of longer sequences, and vice versa.

A language with a high character-to-vocabulary ladderpath index requires greater reconstruction effort to build its vocabulary from character-level units, often because word-level units contain richer internal structure such as stems, affixes, and recurring character sequences; once constructed, these word-level units can then serve as larger reusable building blocks at the corpus level, reducing the vocabulary-to-corpus ladderpath index.
From a cognitive perspective, this corresponds to a greater investment in learning and storing a morphologically rich vocabulary in long-term memory, offset by reduced working memory demands during real-time language use, as more of an utterance can be assembled from prefabricated chunks.
Conversely, a language with a low character-to-vocabulary ladderpath index builds simpler words that are less costly to learn, but these smaller building blocks require more assembly steps at the corpus level, increasing the vocabulary-to-corpus ladderpath index and placing greater demands on working memory during real-time processing.
The negative correlation between the two ladderpath indices provides a quantitative expression of this cognitive trade-off: the total reconstructive effort is partitioned between constructing the vocabulary from characters and assembling the corpus from vocabulary items, and the partition differs across languages while the overall reconstruction cost remains comparatively stable.

\subsection*{Ladderons overlap with human-defined linguistic building blocks}

A natural question is to what extent the ladderons identified by the ladderpath approach, without any linguistic input, correspond to words in the natural vocabulary. 
To assess this, we compare the set of ladderons extracted from each corpus at the character level with the natural vocabulary, defined as the set of word types attested at least twice in the corresponding CoNLL-U files, so as to reduce the influence of rare or highly corpus-specific word types.
The ladderpath algorithm operates solely on character sequences and has no access to word boundaries, morphological annotations, or any other linguistic information; any correspondence between ladderons and words is therefore emergent rather than imposed.

\begin{figure}[ht!]
    \centering
    \includegraphics[width=1.0\linewidth]{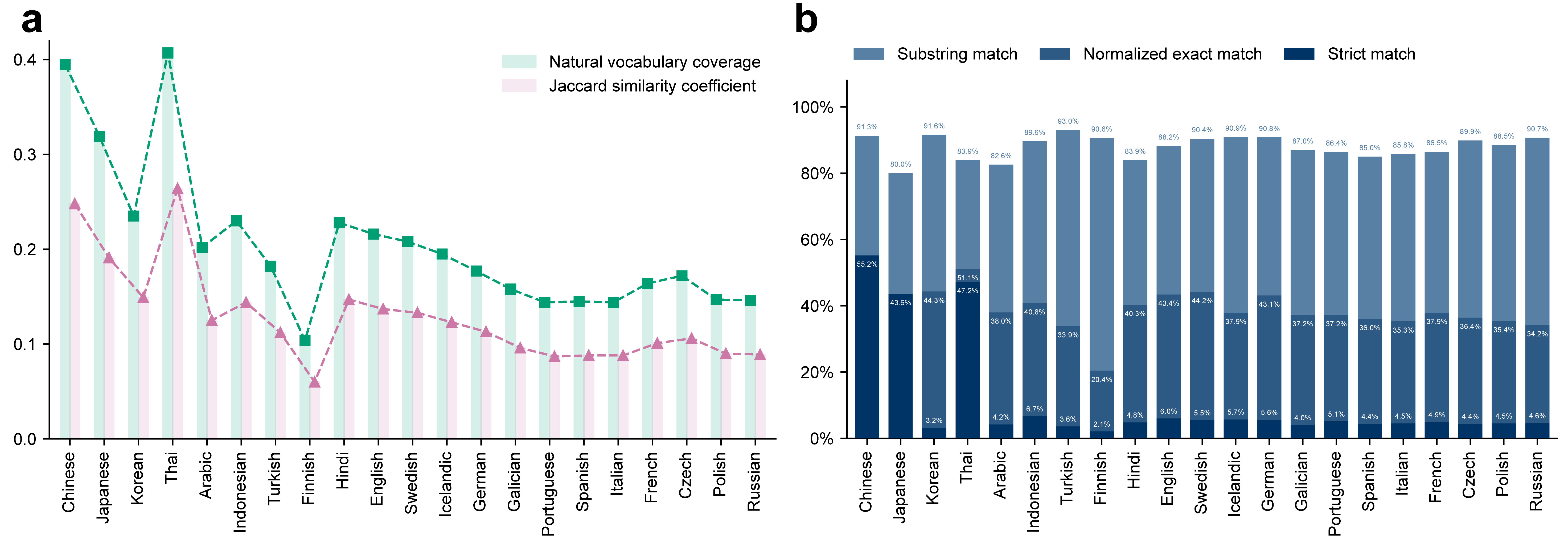}
    \caption{\textbf{Overlap between ladderons and natural vocabulary.}
(a) Overlap between all character-level ladderons and the natural vocabulary across languages, measured by natural vocabulary coverage (intersection divided by the number of natural-vocabulary word types) and by the Jaccard similarity coefficient.
(b) Overlap between the top 1,000 ladderons ranked by multiplicity and the natural vocabulary. Strict matches require exact string identity, normalized exact matches remove boundary spaces and punctuation before matching, and substring matches count ladderons that occur within natural-vocabulary word types.}
    \label{fig4}
\end{figure}

Across the 21 PUD languages, the intersection between the ladderon set and the natural vocabulary covers $(20.6 \pm 1.7)\%$ of natural-vocabulary word types on average, and the Jaccard similarity coefficient between the two sets is $0.128 \pm 0.011$ (Fig. \ref{fig4}a). 
These results indicate a substantial but partial overlap: many natural-vocabulary word types are recovered as ladderons, while the two sets are not identical because ladderons also include sub-word fragments and cross-word sequences that do not coincide with word boundaries.

To examine whether the most important ladderons capture linguistically meaningful structure, we rank ladderons by their multiplicity and analyze the top 1,000 ladderons in each language. 
For these high-ranked ladderons, we consider three types of matches with the natural vocabulary (Fig. \ref{fig4}b). 
A \textit{strict match} requires the raw ladderon string to exactly match a word type in the natural vocabulary, including any surrounding spaces or punctuation. 
A \textit{normalized exact match} first removes such boundary spaces and punctuation from the ladderon and then tests for exact correspondence with a vocabulary item. 
A \textit{substring match} asks whether the ladderon occurs as a substring of any word type in the natural vocabulary. 
Across languages, normalized exact matches account for a substantial fraction of the top-ranked ladderons, mostly around 35--45\%. Strict matches are generally lower, especially in languages where word boundaries are marked by spaces, indicating that many mismatches arise from boundary characters attached to otherwise valid word forms. This is consistent with the relatively high strict-match values observed for languages such as Chinese, Japanese, and Thai, where spacing plays a smaller role in word segmentation. 
When substring matches are considered, the coverage rises above 80\% for every language and reaches up to 93\%, showing that the most frequently reused ladderons are strongly aligned with linguistically meaningful word-level or subword-level units.

Exact matched ladderons, i.e., items identical to attested words, are primarily composed of function words and high-frequency content words.
Function words constitute the most consistent and prevalent matches across all languages. These are typically high-frequency articles, prepositions, conjunctions, pronouns, and auxiliary verbs. Examples include \textit{the}, \textit{of}, \textit{in}, \textit{and}, \textit{to}, \textit{it}, \textit{is} in English; \textit{de}, \textit{en}, \textit{que}, \textit{la}, \textit{el}, \textit{un}, \textit{a} in French; и (and), в (in), не (not), на (on), я (I), быть (be), and что (that) in Russian; \textit{ja} (and), \textit{on} (is), \textit{se} (it), \textit{hän} (he/she), \textit{ei} (not), and \textit{että} (that) in Finnish; and characters such as 的 (of), 在 (at), 了 (le), 和 (and), 是 (be), 中 (in), and 他 (he) in Chinese. In languages such as Japanese and Korean, this category also includes high-frequency particles, for instance, Japanese を (accusative), は (topic), and が (nominative), as well as Korean 이/가 (nominative), 을/를 (accusative), and 은/는 (topic/contrast).
High-frequency content words are nouns, verbs, and adjectives that appear frequently in the corpus and express fundamental concepts (e.g., \textit{man}, \textit{day}, \textit{war}, \textit{time}, \textit{able}, \textit{old}, \textit{new}, \textit{ran}, \textit{go}, \textit{act}, \textit{work}, along with their counterparts in other languages). Given that the PUD corpus is derived from news discourse, domain-specific high-frequency words such as \textit{government}, \textit{company}, \textit{president}, \textit{America}, and \textit{China}—and their corresponding terms in other languages—are also present.
The analysis above reveals a high degree of correspondence between exact matched ladderons and high-frequency word lists from general corpus. This alignment indicates that the unsupervised algorithm successfully captures universal linguistic knowledge.

Unmatched ladderons are primarily composed of three parts: syllabic/alphabetic fragments, sub-word fragments, and cross-word sequences.
Syllabic/alphabetic fragments constitute the predominant component. This category includes individual letters that do not form words in languages like English, French and German, Chinese characters used for transliterating loanwords (e.g., 斯, 尔, 尼), syllabic combinations (e.g., \T{nt}, \T{ig}, \T{ord}, \T{ical} in English; Katakana in Japanese used for loanwords, such as ラン \T{-ran/-run}, スト \T{st-/-st}, and ット \T{-t/-tte}), and year prefixes (e.g., \T{201-}, \T{200-}, \T{19-}). Although highly reused in the ladderpath construction process, these fragments do not exhibit a stable correspondence with specific grammatical or semantic functions and thus do not constitute entries in the mental lexicon.
Excluding these syllabic/alphabetic fragments, the remaining sub-word fragments and cross-word sequences reveal the ladderpath algorithm's capacity to capture linguistic features beyond the word level.
Sub-word fragments are predominantly high-frequency morphemes, particularly inflectional and derivational affixes with significant grammatical functions, which are typically not treated as separate units in corpus tokenization. Examples include English \textit{-ing}, \T{-ed}, \T{-tion}, \T{-ly}, \T{-er}, \T{re-}; Spanish \T{-ción / -sión} (noun suffixes), \T{-mente} (adverbial suffix), and \T{-ado} (past participle); Russian -ов (noun plural genitive), -ен (adjectival/participial), and -ическ (adjectival suffix); and Finnish \T{-ssa} (inessive case), \T{-sta} (elative case), \T{-lla} (adessive case), and \T{-n} (genitive case); and bound morphemes in Chinese 性 (property/nature), 业 (industry/enterprise), 意 (meaning/sense), and 政 (government/politics).
Cross-word sequences include high-frequency chunks such as \T{of the}, \T{in the}, \T{the first}, \T{has been}, \T{to be}, \T{part of the}, and \T{according to} in English; \T{de la}, \T{en el}, \T{y la}, \T{Mientras tanto}, \T{se convirtió en}, and \T{que se} in Spanish; and 一个, 之间的, and 的一 in Chinese. Regardless of whether they convey a complete meaning, these lexical bundles can serve as frames for the construction of phrases or sentences. This category also encompasses combinations of words and punctuation, which can be interpreted as a form of chunking that merges discourse elements with lexical items. Examples include Chinese \T{，但} (comma + \T{but}, adversative), \T{后，} (\T{after} + comma, succession), and \T{了。} (sentence-final particle + period); Japanese \T{した。} (past tense + period, polite style), \T{は、} (topic marker + comma), and \T{である。} (\T{is} + period, formal style); and English \T{, but} and \T{, and}.

These results suggest that the ladderpath algorithm, despite operating without any linguistic knowledge, identifies structures that are not arbitrary statistical fragments. Instead, the high-ranked ladderons consistently correspond to functional linguistic units, including words, stems, affixes, and lexical bundles.   
This convergence between a purely algorithmic decomposition and the vocabulary attested in linguistically annotated corpora provides independent support for interpreting ladderons as analogues of cognitive chunks. In essence, the reusable substructures that the algorithm deems most efficient for reconstruction are largely the same units that languages have conventionalized.

\section*{Discussion}
\label{S:3}

In this study, we introduced the ladderpath index as a measure of language complexity grounded in algorithmic information theory. 
We apply the ladderpath approach to 21 parallel corpora of the PUD dataset and to an additional set of parallel texts translated by an artificial intelligence (AI) model (DeepSeek), at both the character level and the bit level.
This analysis provided evidence for two longstanding hypotheses in quantitative linguistics.
First, the near-invariance of the ladderpath index across languages supports the equi-complexity hypothesis that the reconstruction complexity of the 21 corpora is approximately constant.
Second, the negative correlations between character inventory size and corpus length, and between the character-to-vocabulary and vocabulary-to-corpus ladderpath indices, provide quantitative evidence for the trade-off hypothesis that total complexity is conserved across linguistic levels, with different languages distributing reconstruction effort differently between vocabulary reconstruction and corpus assembly.
Furthermore, we showed that the ladderons identified by the algorithm, without access to any linguistic information, substantially overlap with the words and sub-word units of the natural vocabulary.
The top-ranked ladderons by multiplicity closely resemble words, stems, and affixes rather than arbitrary character fragments, suggesting that the ladderpath algorithm independently recovers units that languages conventionally employ as morphological and lexical building blocks.
These findings connect naturally to cognitive models of hierarchical chunking in language processing, where the balance between long-term memory storage and real-time working memory assembly is subject to shared constraints across all human languages.

The ladderpath index captures the complexity of reconstructing a corpus through hierarchical reuse of repeated substructures. 
It can be understood as a specific, computable facet of Kolmogorov complexity, one that is restricted to savings achievable through concatenation and reuse rather than through arbitrary computable transformations. 
It does not capture all aspects of language complexity. Syntactic dependencies, semantic ambiguity, pragmatic inference, and other dimensions of linguistic structure that do not manifest as repeated character or word sequences are not reflected in the ladderpath index.
The near-invariance we observe therefore pertains specifically to the repetitive-compositional dimension of complexity. 
The ladderpath index also differs from Shannon entropy, which characterizes the average uncertainty per symbol given a probability distribution over an ensemble of messages. 
Entropy is a statistical property of an ensemble and requires distributional assumptions, whereas the ladderpath index is defined for individual objects and is deterministic. 
The two measures may correlate in practice—corpora with higher entropy may tend to have higher ladderpath indices, but they address different questions and are not interchangeable.

The present study is limited by the scope of the datasets used. The PUD dataset contains 21 languages represented by 1,000 sentences each, and the supplementary AI-translated dataset contains additional parallel texts translated into the same target languages. While these datasets span several language families and writing systems, they still represent only a small fraction of the world's linguistic diversity. Furthermore, both datasets are based on translated texts: the PUD sentences are translations from news and Wikipedia pages, whereas the supplementary dataset was generated through machine translation from Chinese and English news-style source texts. Translated text is known to exhibit interference effects from the source language, including unusual word order, atypical lexical choices, and reduced stylistic diversity, and machine-translated text may introduce additional regularities or artifacts that differ from naturally produced language.

The overlap between ladderons and natural vocabulary words suggests potential applications in natural language processing (NLP), particularly in tokenization. 
Current subword tokenization methods such as Byte Pair Encoding (BPE) and WordPiece construct vocabularies by iteratively merging frequent character pairs, optimizing for compression of a training corpus \cite{Sennrich2016, Kudo2018}. 
The ladderpath approach offers a principled alternative grounded in algorithmic information theory: rather than greedily merging frequent pairs, it identifies the set of reusable substructures that minimizes the total reconstruction cost of a corpus.
The fact that high-ranked ladderons already recover a substantial proportion of linguistically meaningful words and morphemes, without any linguistic supervision, suggests that ladderpath-based tokenization could produce subword vocabularies that are both efficient and linguistically interpretable. Whether this translates into improved performance on downstream tasks such as machine translation, language modeling, or morphological analysis requires further investigation.

\section*{Methods and Materials}
\label{S:4}

\subsection*{Corpora}

We compare language complexities through analyzing the Parallel Universal Dependencies (PUD) dataset \cite{Zeman2017, deMarneffe2021}. 
The PUD dataset which contains 1,000 sentences were randomly selected from online news or Wikipedia articles, with 750 originally in English, 100 in German, 50 in French, 50 in Italian and 50 in Spanish.
The 1,000 sentences were translated by professional translators to other 20 languages, resulting in 21 corpora of 1,000 parallel sentences.
The 21 corpora are in Arabic, Chinese, Czech, English, Finnish, French, Galician, German, Hindi, Icelandic, Indonesian, Italian, Japanese, Korean, Polish, Portuguese, Russian, Spanish, Swedish, Thai and Turkish, respectively.
Each corpus is stored in a CoNLL-U formatted file \cite{Zeman2017}.
In the CoNLL-U formatted files, each sentence is segmented into word-level tokens using methods that vary according to the properties of the language and its writing system.
We define the word-level tokens in the CoNLL-U formatted file of a corpus as the \textit{natural vocabulary} of the corresponding language.
In addition to the PUD dataset, we also constructed a supplementary dataset consisting of AI-translated news corpora. See SI Section S2 for more details.

\subsection*{Experiments}

Analyzing a corpus using the ladderpath approach requires identifying the granularity at which the ladderpath index and the size index are computed.
A {\em basic building block} is the smallest non-separable structure at a given level of analysis.
In this work, a basic building block can be a character or symbol in a given language, a word in the natural vocabulary of a language, or a binary symbol 0 or 1. 

We apply the ladderpath approach to the corpora in three configurations. 
In the first configuration, characters and symbols in a language are basic building blocks.
We treat each corpus as a sequence of characters and compute the ladderpath index directly from the corpus to the character level. 
The size index in this case corresponds to the number of characters or symbols in the corpus. 
This provides a baseline measurement of the structural complexity of each corpus at the finest granularity available in the natural language representation.

In the second configuration, we encode each corpus in the Unicode Transformation Format-8 (UTF-8), which represents every character as a sequence of one to four bytes. 
This encoding converts each corpus into a binary representation. 
We then apply the ladderpath approach to the UTF-8 version of each corpus, treating bits as the basic building blocks, and compute the corresponding ladderpath index and size index. 
This configuration allows us to examine corpus structure at the bit level, where the basic building block is uniform across all languages regardless of their native writing systems.

In the third configuration, we decompose the analysis into two stages through the natural vocabulary of each language (Fig. \ref{fig3}c). In the first stage, we treat characters and symbols as the basic building blocks and compute the ladderpath index required to reconstruct the natural vocabulary; this gives the character-to-vocabulary cost $A$. In the second stage, we treat the words in the natural vocabulary as the basic building blocks and compute the ladderpath index required to reconstruct the corpus; this gives the vocabulary-to-corpus cost $B$; see SI Section S3 for more details. The two ladderpath indices are computed separately but are linked by the same prior natural vocabulary. We then compare their combined cost, $A+B$, with the direct character-level ladderpath index $\lambda$ computed in the first configuration, where the corpus is reconstructed directly from characters.

\section*{Data availability}
The data and source code for the experiments and ladderpath calculations is publicly available at 
https://github.com/yuernestliu/LanguageComplexity\_ladderpath.
Supplementary Information can also be found in the repository.

\section*{Competing interest}
The authors declare no competing interests.

\section*{Acknowledgments}
R.L. was supported by the start-up fund for scientific research of BNUZ.
P.L. was supported by the start-up fund of the University of Rhode Island.
Y.L. was supported by the National Natural Science Foundation of China (Grant No. 12205012) and Basic and Applied Basic Research Foundation of Guangdong Province (Grant No. 2025A1515012923).

\bibliographystyle{unsrtnat}
\bibliography{ref.bib}


\end{document}